# Spectral Simplicial Theory for Feature Selection and Applications to Genomics

Kiya W. Govek*, Venkata S. Yamajala*, and Pablo G. Camara[#]

Department of Genetics and Institute for Biomedical Informatics,

Perelman School of Medicine, University of Pennsylvania,

3400 Civic Center Blvd., Philadelphia, PA 19104.

* These authors contributed equally to this work.

[#] Correspondence to: pcamara@pennmedicine.upenn.edu

**Abstract:** The scale and complexity of modern data sets and the limitations associated with testing large numbers of hypotheses underline the need for feature selection methods. Spectral techniques rank features according to their degree of consistency with an underlying metric structure, but their current graph-based formulation restricts their applicability to point features. We extend spectral methods for feature selection to abstract simplicial complexes and present a general framework which can be applied to 2-point and higher-order features. Combinatorial Laplacian scores take into account the topology spanned by the data and reduce to the ordinary Laplacian score in the case of point features. We demonstrate the utility of spectral simplicial methods for feature selection with several examples of application to the analysis of gene expression and multi-modal genomic data. Our results provide a unifying perspective on topological data analysis and manifold learning approaches.

**Significance Statement:** Manifold learning methods have emerged as a way of analyzing the large high-dimensional data sets that are currently generated in many areas of science. They assume the data has been sampled from an unknown manifold which is approximated with a graph, and utilize spectral graph techniques to perform unsupervised feature selection and dimensionality reduction. However, graphs provide only partial approximations to manifolds, precluding the application to features with a complex combinatorial structure. Relatedly, these methods cannot take into account the manifold's topology. In this work, we extend spectral methods for feature selection to topological spaces built from data and present a general framework for feature selection. We present specific applications to the analysis of gene expression and multi-modal genomic data.

**Introduction**

A classical problem in statistics and data analysis is the identification of features that discriminate samples from discrete and identifiable classes. During the $20^{th}$ century, numerous non-parametric tests were developed to that end, including the Kolmogorov-Smirnov, Wilcoxon, and Mann-Whitney tests. These tests have become ubiquitous across all areas of science. In these problems, classes are inherent to the formulation of the problem or emerge from an underlying metric or semi-metric structure on the data which permits grouping samples into clusters and subsequently identifying differential features. However, there are also many situations in science where samples cannot be naturally arranged into discrete classes, for instance when they are drawn from a continuous process. In those situations, common tests of significance cannot be applied and feature selection becomes cumbersome.

When samples cannot be arranged into classes, a standard approach to feature selection is to use variance as a proxy (1). This approach assumes that features that have high variance across the entire data set are more informative about the structure of the data than features that take a constant value. Although the use of variance for feature selection has offered many benefits in multiple domains, it only makes use of the set structure and ignores the richer metric or semi-metric structure of the data when it is available (Fig. 1). In this spirit, *manifold learning* methods, such as Laplacian score (2) and diffusion maps (3), have become increasingly popular in the past decade, as technological progress has allowed generating the large amounts of high-dimensional vector-like data they require. The assumption of these methods is that each vector of measurements specifies the coordinates of a point in a high-dimensional manifold. A nearest neighbor graph is then constructed to approximate the unknown manifold from the data and spectral graph techniques are utilized for unsupervised feature selection and dimensionality reduction.

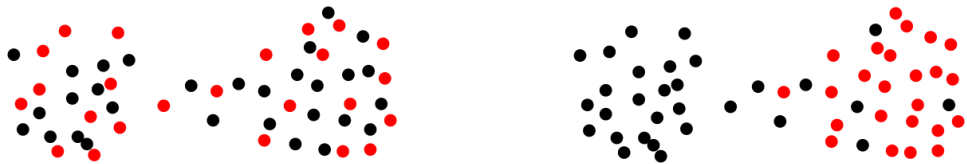

*Figure 1. Variance can be used as a proxy for unsupervised feature selection but does not take into account the underlying metric structure of the data. In the figure a finite metric space is labeled according to two binary features with equal variance. However, only the feature on the right shows a large degree of consistency with the metric space. Manifold learning approaches to feature selection prioritize features according to the degree of consistency with the underlying metric structure of the data.*

Despite their utility and widespread use, graphs provide only partial approximations to manifolds, as they do not capture 3-point and higher-order relations. A more complete approximation is provided by simplicial complexes (4), algebraic generalizations of graphs that, apart from vertices and edges, can include higher-dimensional polytopes such as triangles, tetrahedrons, etc. They are a central concept in algebraic topology, geometry, and combinatorics, where they provide a general framework for the description and computation of topological spaces. These spaces include manifolds as well as more general spaces like topologically stratified spaces, where the dimensionality of the space can change across the space.

In this paper, we adapt the use of spectral techniques to simplicial complexes and introduce a general framework for unsupervised feature selection which extends current graph-based methods. To that end, we introduce the combinatorial Laplacian score for feature selection in simplicial complexes built from the data (Fig. 2). Although our framework is applicable to all areas of data analysis, we present several specific applications to genomics, where spectral methods allow for new types of analyses. We also briefly consider the closely related problem of feature extraction, where synthetic features that are maximally informative are generated, and provide a generalization of Laplacian Eigenmaps (5). Our work contributes to bridging the gap between manifold learning and topological data analysis by extending some of the foremost tools of manifold learning into abstract simplicial complexes.

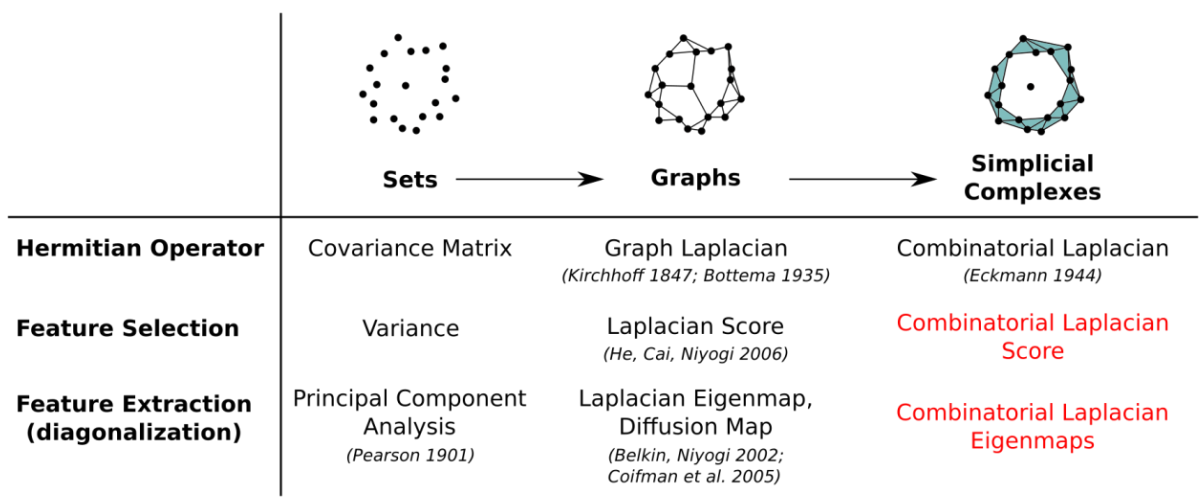

*Figure 2. Summary of various related approaches to unsupervised feature selection and extraction.*

**Geometric Feature Selection**

The main objects of study in this paper are a finite data set (often also termed as point cloud) $X$ with a notion of distance or dissimilarity $\|x_i - x_j\|$ and a set of features $\{f_r\}$ defined as maps from $X$ into a formally real field $\mathbb{F}$. In 2006, He, Cai, and Niyogi proposed an algorithm for unsupervised feature selection called Laplacian score (2). They construct a weighted nearest neighbor graph $G$ with nodes $X$ and adjacency matrix $A$ and introduce an inner product in $G$ defined by

$$\langle f_r, f_s \rangle_G = \sum_{x_i \in X} w(x_i)\, f_r(x_i)\, f_s(x_i)$$

with $w(x_i) = \sum_j A_{ij}$. The Laplacian score of the $r$-th feature is then given by the Rayleigh quotient of the normalized graph Laplacian with respect to this inner product,

$$R_r = \frac{\langle \tilde{f}_r, L_G \tilde{f}_r \rangle_G}{\langle \tilde{f}_r, \tilde{f}_r \rangle_G}$$

where

$$L_G = I - \text{diag}(w)^{-1} A$$

$$\tilde{f}_r = f_r - \frac{\langle f_r, \mathbf{1} \rangle_G}{\langle \mathbf{1}, \mathbf{1} \rangle_G}$$

and $\mathbf{1}(x_i) = 1, \forall x_i \in X$, is the unit feature vector. The Laplacian score ranks features according to their consistency with the structure of $G$. Specifically, features with small values for $R_r$ take high values in highly connected nodes of $G$. This approach to unsupervised feature selection has become widespread, as it offers a substantial statistical power compared to ranking features according to their variance (2). In what follows, we generalize these notions to simplicial complex representations of the data.

**Preliminary Definitions.** We first recall some standard definitions from algebraic topology that will be used below (4). We define an ordered abstract simplicial complex $K$ on a finite set $V$ as a collection of ordered subsets of $V$ which is closed under inclusion, i.e. $\tau \subset \sigma \in K \Rightarrow \tau \in K, \forall \sigma$. The $(q+1)$-dimensional elements of $K$ are called $q$-simplices. We denote the set of $q$-simplices of $K$ by $S_q(K)$.

There are multiple ways to construct an ordered abstract simplicial complex from a data set $X$ and an order relation $x_0 < x_1 < \cdots < x_N$ among the elements of $X$ (6). The Čech complex $C(X, \varepsilon)$ is defined as the abstract simplicial complex that results from considering the set of open balls with radius $\varepsilon \in F^+$ centered at the elements of $X$, where balls correspond to 0-simplices, pairwise intersections to 1-simplices, triple intersections to 2-simplices, etc. (Fig 3). The utility of the Čech complex is ensured by the nerve theorem: if the elements of $X$ were sampled from a topological space $\mathcal{M}$, under certain sampling assumptions the Čech complex approximates the topology of $\mathcal{M}$. In particular, the 1-skeleton (i.e. the set of 0- and 1-simplices) of the Čech complex is a graph of the type considered in manifold learning approaches. Since computing triple and higher-order intersections among balls is computationally costly, the Vietoris-Rips complex (defined as the clique complex of the 1-skeleton of a Čech complex) is often used in practice as an approximation to the Čech complex (6). Additionally, Uniform Manifold Approximation and Projection (UMAP) (7) and Mapper (8) can be used to generate reduced abstract simplicial complex representations of the data.

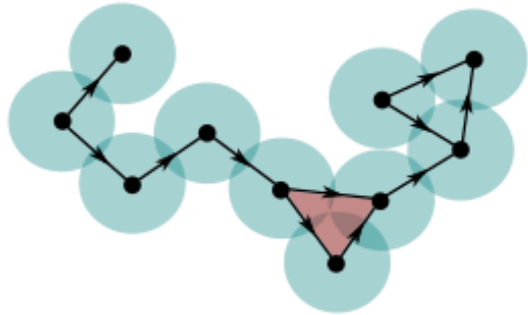

Figure 2. Čech complex constructed from intersections of fixed-radius balls centered at the elements of the point cloud $X$. In this example, the elements of $X$ are ordered according to the horizontal coordinate.

Given a $(q + 1)$-simplex $\sigma = \{v_0, \ldots, v_{q+1}\} \in K$, we define its boundary as the linear combination of $q$-simplices

$$\partial \sigma = \sum_{k=0}^{q+1} (-1)^k \{v_0, \ldots, v_{k-1}, v_{k+1}, \ldots, v_{q+1}\}$$

where 1 is the unit element of F. More generally, the boundary operator can act on linear combinations of $q$-simplices. We denote by $\text{sgn}(\tau, \partial\sigma)$ the sign of a $q$-simplex $\tau$ contained in the boundary of $\sigma$.

We now turn our attention to maps from $X$ into F. We define a $q$-point feature of $X$ as a map $f\left(x_{i_1}, \ldots, x_{i_q}\right)$ from $X^q$ into F. For instance, if the elements of $X$ are trading agents, the frequency two agents enter into a commercial transaction to trade a specific type of asset would constitute a 2-point feature of $X$. Every $q$-point feature of $X$ induces a map from the $(q-1)$-simplices of the Čech complex $C(X, \varepsilon)$ into F in the obvious way. More generally, we define a discrete $q$-form or $q$-cochain $f^{(q)}$ on a simplicial complex as a map from $S_q \subset K$ into F. If the simplicial complex is obtained by discretizing a manifold, for instance through triangulation, discrete $q$-forms on $K$ approximate differential $q$-forms on the manifold (9). In the case of reduced abstract simplicial complex representations, such as those produced by Mapper, 1-point features can also induce $q$-forms with $q > 1$.

With these definitions we can extend the inner product $\langle , \rangle_G$ to a weighted inner product between $q$-forms on simplicial complexes

$$\langle f_r^{(q)}, f_s^{(q)} \rangle_K = \sum_{\tau \in S_q} w(\tau)\, f_r^{(q)}(\tau)\, f_s^{(q)}(\tau)$$

where $w(\tau) \in F$ is the weight of $\tau$. In particular, notice the inner product between 0-forms is equivalent to $\langle , \rangle_G$ where $G$ is the 1-skeleton of $K$.

Finally, we introduce Eckmann's generalization of the graph Laplacian to discrete $q$-forms on simplicial complexes (10), namely the combinatorial Laplacian on simplicial complexes, defined as

$$L_K^{(q)} = L_K^{(q),\uparrow} + L_K^{(q),\downarrow}$$

$$(L_K^{(q),\uparrow} f_r^{(q)})(\tau) = \sum_{\substack{\sigma \in S_{q+1}: \\ \tau \in \partial \sigma}} \frac{w(\sigma)}{w(\tau)} f_r^{(q)}(\tau) + \sum_{\substack{\tau' \in S_q: \tau \neq \tau', \\ \tau, \tau' \in \partial \sigma}} \frac{w(\sigma)}{w(\tau)} \operatorname{sgn}(\tau, \partial \sigma) \operatorname{sgn}(\tau', \partial \sigma) f_r^{(q)}(\tau')$$

$$(L_K^{(q),\downarrow} f_r^{(q)})(\tau) = \sum_{\rho \in \partial \tau} \frac{w(\tau)}{w(\rho)} f_r^{(q)}(\tau) + \sum_{\substack{\tau': \\ \tau \cap \tau' = \rho}} \frac{w(\tau')}{w(\rho)} \operatorname{sgn}(\rho, \partial \tau) \operatorname{sgn}(\rho, \partial \tau') f_r^{(q)}(\tau')$$

In these expressions $L_K^{(q),\uparrow}$ only depends on $S_{q+1}$, whereas $L_K^{(q),\downarrow}$ only depends on $S_{q-1}$. In particular, $L_K^{(0)} = L_K^{(0),\uparrow}$ is independent of the ordering of $X$ and is identical to the graph Laplacian $L_G$, with $G$ the 1-skeleton of $K$. Furthermore, if $K$ is the result of discretizing a manifold, the normalized combinatorial Laplacian approximates the continuous Laplace-Beltrami operator of the manifold (9). With these elements at hand, we are now ready to introduce the combinatorial Laplacian score for $q$-point features.

**Combinatorial Laplacian Score.** Given a simplicial complex representation $K$ of a finite data set and a set of features, we define the combinatorial Laplacian score as

$$R_r^{(q)} = \frac{\langle \tilde{f}_r^{(q)}, L_K^{(q)} \tilde{f}_r^{(q)} \rangle_K}{\langle \tilde{f}_r^{(q)}, \tilde{f}_r^{(q)} \rangle_K}$$

where

$$\tilde{f}_r^{(q)} = f_r^{(q)} - \frac{\langle f_r^{(q)}, \mathbf{1}^{(q)} \rangle_K}{\langle \mathbf{1}^{(q)}, \mathbf{1}^{(q)} \rangle_K}$$

In this expression, $\{f_r^{(q)}\}$ are the discrete $q$-forms on $K$ induced by the set of features, and $\mathbf{1}^{(q)}: S_q \to 1$ denotes the unit $q$-form on $K$. Hence, the combinatorial Laplacian score for 0-forms $R_r^{(0)}$ reduces to the ordinary Laplacian score on the 1-skeleton graph of $K$. For higher-order forms, the combinatorial Laplacian score ranks features according to their degree of localization along homological features of the simplicial complex. For example, $R_r^{(1)}$ takes low values for features that take high values along non-contractible loops of the data (Fig. 4), $R_r^{(2)}$ takes low values for features that take high values along the 2-dimensional walls that line cavities of the data, etc. In summary, the combinatorial Laplacian score extends the ordinary Laplacian score for feature selection to higher-order relations of the data.

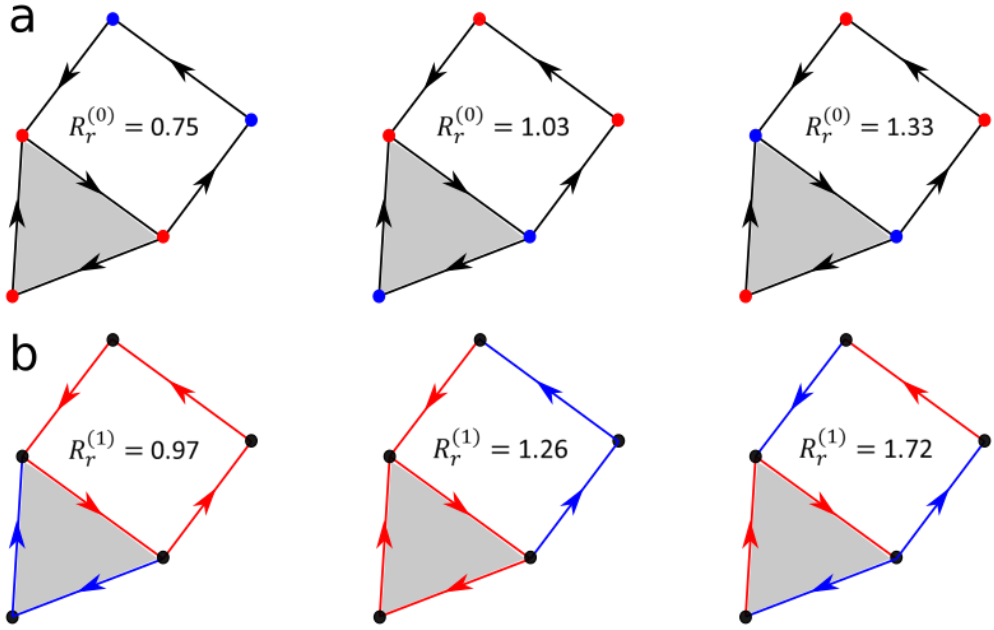

*Figure 4. Combinatorial Laplacian score for a set of binary 1-point (a) and 2-point (b) features on an oriented Čech complex. Features can take values 0 (blue) or 1 (red). Features that take high values on highly connected nodes in the simplicial complex have lower values of $R_r^{(0)}$ than features that take high values on disconnected nodes (a). Analogously, features that take high values on edges that form non-contractible loops in the simplicial complex have lower values of $R_r^{(1)}$ than features that take high values on disconnected or contractible paths (b).*

The statistical significance of the combinatorial Laplacian score can be estimated by randomization of $X$. For each feature, it is possible to build a null-distribution for $R_r^{(q)}$ by randomly permuting the elements of $X$ and computing $R_r^{(q)}$ multiple times. By controlling the rate of type I errors using standard procedures (11, 12), it is then possible to fix the Čech complex parameter $\varepsilon$ (or the parameters of UMAP and Mapper in the case of those representations) such that the number of rejected null hypotheses at a fixed false discovery rate (FDR) is maximized. The process is illustrated in Fig. 5, where we use the combinatorial Laplacian score to identify informative pixels in the MNIST dataset.

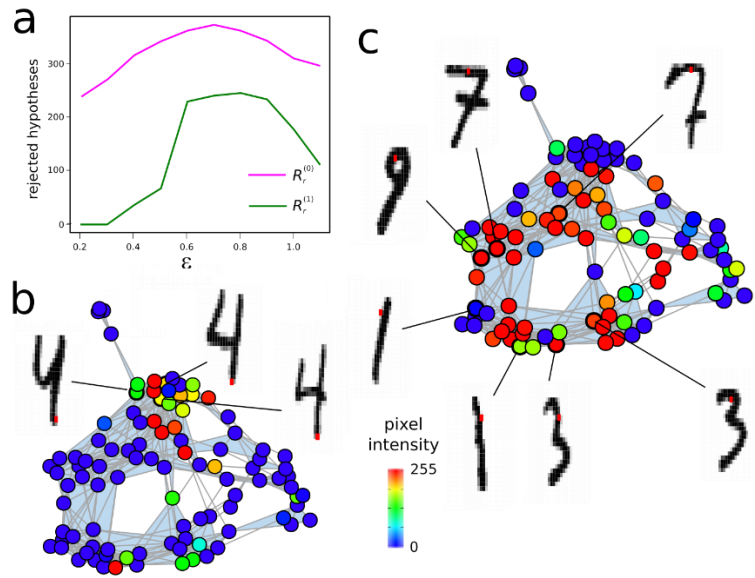

*Figure 5. Feature selection on the MNIST dataset using the combinatorial Laplacian score. Each sample consists of a 28x28 grey-scale image of a hand-written digit from 0 to 9. Each pixel represents a feature. A Vietoris-Rips complex is built using Pearson's correlation distance. The degree of localization of each feature in the simplicial complex is assessed using the combinatorial Laplacian score. (a) Number of rejected null hypothesis as a function of the Vietoris-Rips complex parameter $\varepsilon$ at a FDR of 0.05 (p-values estimated by randomization with 5,000 permutations; FDR controlled using Benjamini-Hochberg procedure). (b) Vietoris-Rips complex ($\varepsilon$ = 0.7) colored by the intensity of a pixel that is significant under $R_r^{(0)}$ (q-value < 0.005) but not under $R_r^{(1)}$ (q-value = 0.5). The intensity of the pixel is high in a densely connected, topologically trivial region of the complex containing images of the digit `4`. Images associated to several nodes are shown for reference, with the pixel highlighted in red. (c) Vietoris-Rips complex colored by the intensity of a pixel that is significant under both $R_r^{(0)}$ (q-value < 0.005) and $R_r^{(1)}$ (q-value < 0.005). The intensity of the pixel is high in a densely connected region that surrounds a large non-contractible cycle of the simplicial complex. The cycle is generated by images that belong to the*

*sequence of digits `7`, `3`, `1`, `9`. Images associated to several nodes along the cycle are shown for reference, with the pixel highlighted in red.*

**Applications to Genomics**

The combinatorial Laplacian score introduced in the previous section generalizes spectral graph methods to account for higher-order relations of the data and has applications in all domains of data analysis. Here, we focus on applications to the field of genomics and show its utility for differential expression analysis and multi-modal data analysis. All of these applications represent novel uses of spectral methods in genomics and allow for qualitatively different results compared to conventional methods.

**Differential Expression Analysis.** A standard task in genomics is to identify genes that are differentially expressed among discrete groups of samples (tissue specimens or individual cells) whose transcriptome has been profiled using expression microarrays or RNA-sequencing. However, in many situations samples cannot be naturally arranged or clustered into discrete groups based on their transcriptome. Cells (and the tissues they form) often respond to stimuli from their cellular environment in a continuous way. For example, immune cells acquire immunosuppressive or pro-inflammatory phenotypes in different contexts and there is a continuum of cellular phenotypes interpolating among these states. Furthermore, the transcriptome of a cell is the result of many concurrent molecular pathways taking place in the cell. In this section, we utilize the combinatorial Laplacian score to identify differentially expressed genes without predefining groups of samples or cells, as opposed to conventional approaches to differential expression analysis where at least two pre-defined groups of samples are required. We expect this approach to be of great utility in studying the transcriptional programs underlying dynamic cellular systems such as those that occur during cell differentiation (13).

We assessed the power of the 0-dimensional combinatorial Laplacian score to discriminate differentially expressed genes as compared to conventional methods. To that end, we followed the same approach as in a recent comparative study of single-cell differential expression analysis (14). We considered three simulated single-cell RNA-seq data sets containing two populations of cells with 10% of the genes being differentially expressed between the populations. These data sets were generated based on three real single-cell RNA-seq data sets (GSE45719, GSE74596, and GSE60749-GPL13112). In our comparisons, we considered three conventional methods for differential expression analysis (DeSeq2 (15), edgeR (16), and MAST (17)), one of which (MAST) is specifically designed for single-cell RNA-seq data. As opposed to the combinatorial Laplacian score, these methods require assigning each cell to a cellular population or cluster of cells and are therefore restricted to situations where such an assignment is possible. Additionally, we considered utilizing variance to discriminate differentially expressed genes in absence of cell-population or cluster assignments. In all cases, the discriminating power of the combinatorial Laplacian score as measured by the area under the receiver-operating characteristic curve (AUC) was comparable to that of conventional methods, despite its broader applicability, and well above that of variance (Fig. 6a).

We next applied the 0-dimensional combinatorial Laplacian score in a case where conventional methods are of limited utility. Specifically, we considered single-cell RNA-seq data of 24,911 T-cells infiltrating lung tumors and adjacent normal tissue (18). T-cells in this example do not form well-defined stable clusters according to their transcriptional profile, as evidenced by a tSNE representation of the data (Fig. 6b). Conventional methods for differential expression analysis do not take into account the instability of clusters and can miss expression patterns that are incompatible with the cluster structure. Ranking genes according to their 0-dimensional combinatorial Laplacian score leads to the identification of differentially expressed genes in this example (Fig. 6b). Among the top ranked genes are genes whose expression pattern is well accounted for by the cluster structure of the original analysis (18), such as *FGFBP2*, which is expressed exclusively and ubiquitously by natural killer cells. However, the cluster structure does not account for the expression pattern of many of the top differentially expressed genes identified by the combinatorial Laplacian score. For instance, among the cytotoxic CD8+ T-cells, we observe distinct subpopulations of cells expressing combinations of Hepatitis A Virus Cellular Receptor 2 (*HAVCR2*), Radical S-Adenosyl Methionine Domain Containing 2 (*RSAD2*), and Granzyme K (*GZMK*). The combinatorial Laplacian score hence reveals more complex and richer patterns of cellular heterogeneity than clustering-based approaches.

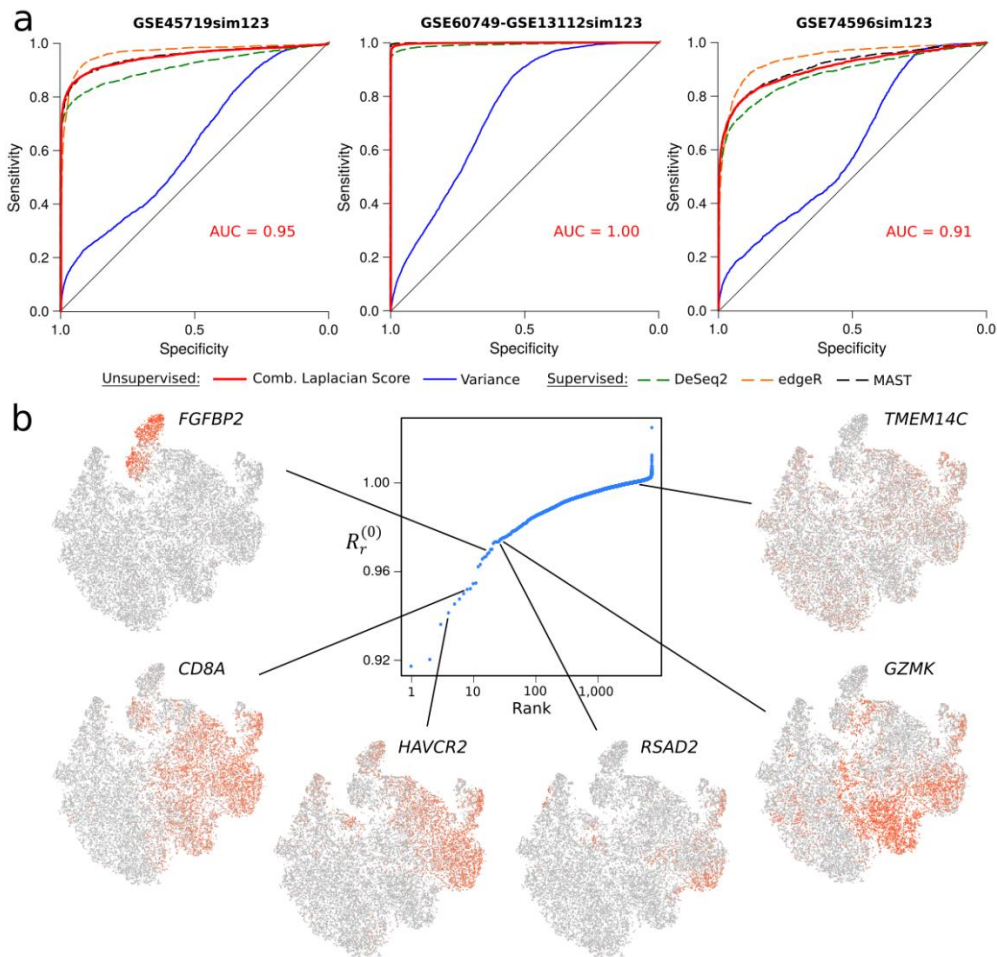

*Figure 6. Differential expression analysis using the combinatorial Laplacian score. (a) ROC curves for three simulated scRNA-seq datasets with 10% of the genes being differentially expressed between two*

*populations of cells. The 0-dimensional combinatorial Laplacian score, DeSeq2, edgeR, MAST, and variance were used to identify differentially expressed genes. Conventional methods (DeSeq2, edgeR, and MAST) in addition took as input a list of labels assigning each cell to the corresponding population. For the analysis with the combinatorial Laplacian score, a Vietoris-Rips complex was built using Pearson's correlation distance. The parameter of the complex was determined in each case by maximization of the number of rejected null-hypotheses. (b) Analysis of scRNA-seq data of 24,911 T-cells infiltrating lung tumors and adjacent tissues. 7,424 genes with expression detected in more than 1% and less than 25% cells were ranked according to their 0-dimensional combinatorial Laplacian score. t-SNE plots color-coded for expression (grey to red) are shown for some of the top differentially expressed genes identified by this method. For reference, the expression of a gene with high combinatorial Laplacian score (TMEM14C) is also displayed. To facilitate interpretation, we use the same t-SNE embedding as in (18).*

**Cyclic Analysis.** The application to differential expression analysis described in the previous section only makes use of the graph structure of the simplicial complex. Higher-dimensional combinatorial Laplacian scores can be used to disaggregate differentially expressed genes according to their degree of localization along topological features of the data. For example, in dynamic cellular systems involving convergent lineages, strong cell-cycle effects, or complex differentiation paths, cells may span loops in the expression space. In those situations, the 1-dimensional combinatorial Laplacian score permits identifying genes that are differentially expressed along the loops.

To show the utility of the 1-dimensional combinatorial Laplacian score for differential expression analysis, and cyclic analysis in general, we considered single-cell RNA-seq data of the *in vitro* differentiation of mouse embryonic stem cells into motor neurons using two different protocols (19). In these experiments, the starting and final transcriptional states are the same between the two differentiation protocols. However, the intermediate transcriptional states are different in each protocol, leading to branching trajectories in the expression space that subsequently converge into the same final state. We ranked the genes in this example according to their 0- and 1-dimensional combinatorial Laplacian scores on a Vietoris-Rips complex constructed from the top 50 principal components of the data (Fig. 7). Genes having a small value of the 1-dimensional combinatorial Laplacian score were differentially expressed along the two alternative paths of differentiation. These included genes, such as *Lhx3* and *Plk3*, which are known to be upregulated in the intermediate states of one of the two differentiation protocols (Fig. 7). On the other hand, genes having a small value of the 0-dimensional combinatorial Laplacian score but not of the 1-dimensional combinatorial Laplacian score were differentially expressed in the initial or final states of the differentiation process (Fig. 7). Thus, when the expression space has a complex topology, different combinatorial Laplacian scores allow to identify and stratify differentially expressed genes according to the topology of the expression space without pre-defining cellular populations.

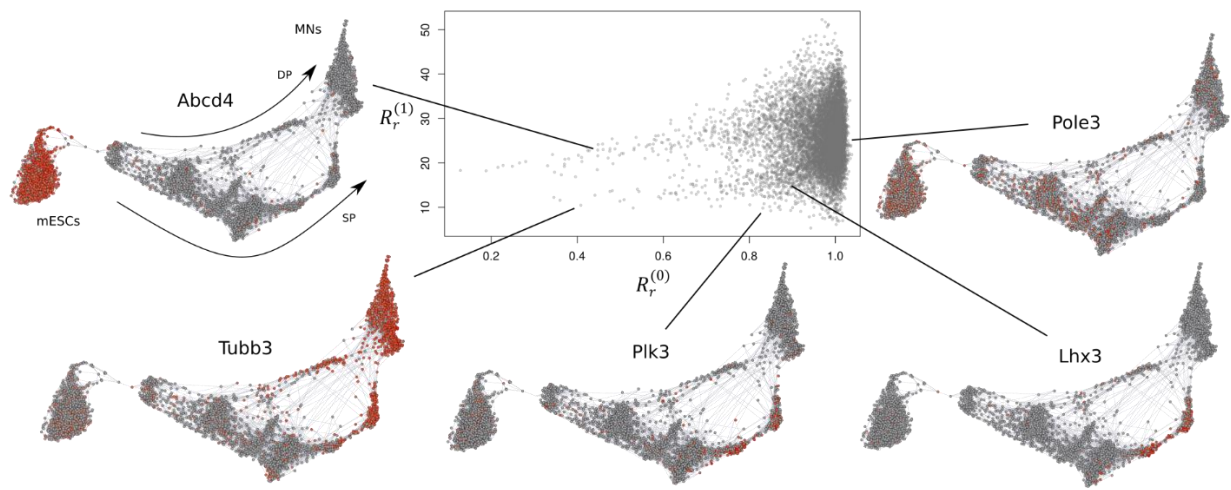

*Figure 7. Differential expression analysis using the 1-dimensional combinatorial Laplacian score. The 0- and 1-dimensional combinatorial Laplacian scores were run over the scRNA-seq expression data of 3,582 cells from the in-vitro differentiation of mouse embryonic stem cells (mESCs) into motor neurons (MNs) using the standard (SP) and direct programming (DP) protocols. The scatter plot represents the 0- and 1-dimensional combinatorial Laplacian scores of 10,063 genes with expression detected in more than 5% and less than 50% cells. A k-nearest neighbor graph (k=4) color-coded for expression (grey to red) is*

*shown for some of the top differentially expressed genes identified by this method. Genes with low value of $\mathrm{R}_r^{(1)}$ are differentially expressed along the cycle spanned by the two alternative paths for differentiation. For reference, the expression of a gene with high 0- and 1- combinatorial Laplacian scores (Pole3) is also displayed. Statistical significance was estimated by randomization with 5,000 permutations. A force-directed layout was used to visualize the k-nearest neighbor graph.*

**Multi-modal Data Analysis.** Spectral methods rank features according to their degree of consistency with a given metric or semi-metric structure, as we have described. However, the data type of the features may differ from that of the metric structure. Spectral methods can therefore be used to perform multi-modal analyses, where features of a given type are ranked according to their degree of consistency with a metric or semi-metric structure of a different type. To show the utility of this type of analysis, we considered two specific applications to multi-modal genomic data.

Spatially-resolved transcriptomics allows to measure the expression level of genes in their spatial context, often with single-cell resolution. We considered single-molecule fluorescence *in-situ* hybridization data of the murine somatosensory cortex (20) and utilized the combinatorial Laplacian score to rank and disaggregate genes according to their spatial expression pattern (Fig. 8a). Genes with low values of the 0-dimensional Laplacian score were differentially expressed across the spatial directions, often being expressed in only a subset of the cortical layers.

Multi-modal data analysis is also important in quantitative trait association studies. These studies seek to establish associations between quantitative phenotypes (e.g. gene expression levels) and genotypes. To show the utility of the combinatorial Laplacian score to establish associations with complex high-dimensional phenotypic spaces, we analyzed somatic mutation data, obtained by whole-exome sequencing, and mRNA expression data of a cohort of 667 low-grade glioma and glioblastoma tumors (21). We used the 0-dimensional combinatorial Laplacian score to identify non-synonymous mutations that are associated with consistent global expression patterns (Fig. 8b). Among the mutated genes with significant 0-dimensional Laplacian score (Benjamini-Hochberg adjusted *q*-value ≤ 0.01; *n* = 1,000 permutations), we observed numerous known cancer-associated genes in adult brain glioma. The significance of this observation was confirmed by a gene-set enrichment analysis using the recurrently-mutated genes reported by MutSig2CV in the same cohort (GSEA *p*-value < 10$^{-4}$).

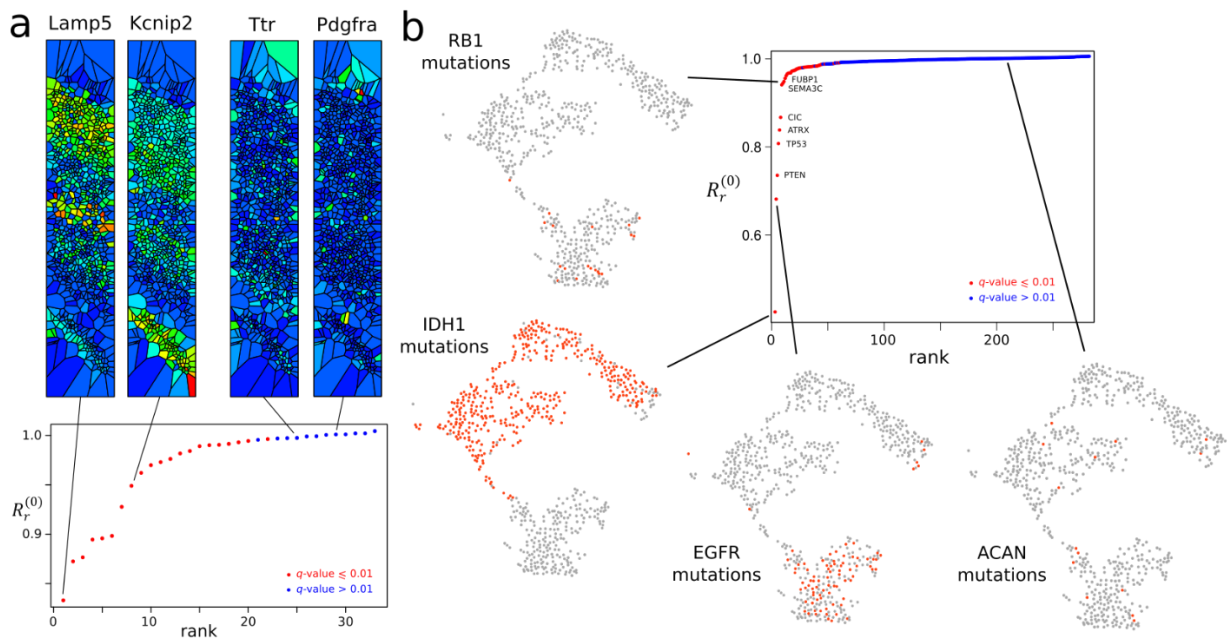

*Figure 8. Analysis of multi-modal genomic data using the combinatorial Laplacian score. (a) The 0-dimensional Laplacian score was utilized to identify genes that are differentially expressed across spatial directions in a section of the murine somatosensory cortex using single-molecule fluorescence in-situ hybridization data. Euclidean distance across the spatial dimensions was used to build a Vietoris-Rips complex and gene expression levels were evaluated using the combinatorial Laplacian score. Genes with low $R_r^{(0)}$ have significant spatial patterns of expression. For reference the expression patterns of two genes with low $R_r^{(0)}$ (Lamp5 and Kcnip2) and two genes with high $R_r^{(0)}$ (Ttr and Pdgfra) are shown. Cells are represented by means of a Voronoi tessellation and are color-coded according to the expression level of the gene (blue: low; red: high). (b) Analysis of whole exome and mRNA sequencing data of 667 low-grade glioma and glioblastoma tumors of The Cancer Genome Atlas using the 0-dimensional Laplacian*

*score. A Vietoris-Rips complex was built using correlation distance among the gene expression patterns. Binary vectors indicating whether a gene is non-synonymously mutated or not were taken as features. The 0-dimensional Laplacian score identifies genes for which their mutation is associated with consistent global expression patterns. A UMAP representation based on the expression data is shown for some representative genes with low (IDH1, EGFR, and RB1) or high (ACAN) value of $R_r^{(0)}$, color-coded according to the somatic mutation status of the gene (orange: non-synonymously mutated; grey: non-mutated or synonymously mutated).*

**Relation to Feature Extraction**

We now turn our attention to feature extraction, a closely related problem to feature selection. In feature extraction, a finite set of synthetic features that optimally capture the structure of the point cloud is engineered. These synthetic features can be then used for dimensionality reduction and de-noising. The Laplacian score of He, Cai, and Niyogi follows from the Laplacian Eigenmaps for dimensionality reduction (5). In what follows, we show how this relation can be naturally extended to simplicial complexes and combinatorial Laplacian Eigenmaps.

To that end, we consider the diagonalization problem of the combinatorial Laplacian,

$$\left(L_K^{(q)} y_i^{(q)}\right)(\tau) = \lambda_i^{(q)} y_i^{(q)}(\tau)$$

Using the Courant-Fischer-Weyl min-max principle we can express the eigenvalues of the combinatorial Laplacian in terms of a feature optimization problem with respect to the combinatorial Laplacian score,

$$\lambda_i^{(q)} = \max_{\substack{U: \\ \dim(U)=\dim(S_q)-i+1}} \left\{ \min_{\tilde{f}_r^{(q)} \in U} R_r^{(q)} \right\}$$

From this relation we observe that the eigenvectors of the combinatorial Laplacian minimize the combinatorial Laplacian score successively across maximal orthogonal directions. In particular, the combinatorial Laplacian score is bounded by the smallest and largest eigenvalues of the combinatorial Laplacian, $\lambda_1^{(q)} \leq R_r^{(q)} \leq \lambda_{\dim(S_q)}^{(q)}$.

We define the $m$-dimensional combinatorial Laplacian Eigenmap for $q$-simplices as the map from $S_q$ into $\mathbb{R}^m$ given by

$$\begin{aligned} \mathcal{M}^{(q)} \colon S_q &\to \mathbb{R}^m \\ \tau &\mapsto \left(y_1^{(q)}(\tau), \ldots, y_m^{(q)}(\tau)\right) \end{aligned}$$

In particular, $\mathcal{M}^{(0)}$ reduces to the ordinary Laplacian Eigenmap in the 1-skeleton of $K$. The combinatorial Laplacian Eigenmap for $q$-simplices thus provide locally-preserving low-dimensional representations of the data where each point in the reduced representation corresponds to $q + 1$ related points in $X$.

**Software**

We have implemented the computation of the 0- and 1-dimensional combinatorial Laplacian score in the open-source R package `RayleighSelection` (https://github.com/CamaraLab/RayleighSelection). To improve performance, we have used parallelization and embedded C++ code for the computationally demanding parts. Additionally, we have included routines for the generation and visualization of simplicial complexes. In its current implementation, the algorithm takes approximately 3 hours to compute the 0-dimensional combinatorial Laplacian score for 10,000 features in a Vietoris-Rips complex with 2,000 vertices and 1 million edges in a standard desktop computer with 8 cores (Table 1). The computation of the 1-dimensional combinatorial Laplacian score is more demanding, and usually requires subsampling the data set or using a high performance computing platform (Table 1).

|  | Vertices | Edges | 2-Simplexes | Time |
|---|---|---|---|---|
| $R_r^{(0)}$ | 1,000 | 249,500 | - | 12m 43s |
|  | 1,500 | 561,750 | - | 59m 18s |
|  | 2,000 | 999,000 | - | 3hr 13m 46s |
| $R_r^{(1)}$ | 100 | 600 | 34,045 | 41m 14s |
|  | 150 | 2,450 | 112,175 | 7hr 19m 6s |
|  | 200 | 5,550 | 268,758 | 31hr 48m 28s |

Table 1. Running times of the 0- and 1-dimensional combinatorial Laplacian scores for several simplicial complexes of different size using the code `RayleighSelection` on a standard 8-core desktop computer.

**Summary and Discussion**

Modern data often comes in the form of large finite metric or semi-metric spaces. In those cases, a widespread analytic approach involves clustering the data points based on the underlying metric structure and using some statistical test to identify features that differ across clusters. However, there are multiple situations where data cannot be naturally structured into clusters and, in the cases where clustering is possible, there are always basic trade-offs inherent to the clustering problem (22). In addition, in this two-step approach the uncertainty in the clustering step is not taken into account in the statistical test used to identify features, leading to potentially misleading conclusions. Spectral methods bypass the clustering step and directly use the underlying metric structure of the data to perform feature selection. However, current methods are graph-based and limited to point features of the data. In this work we have generalized these methods and presented a general framework for feature selection based on abstract simplicial complexes built from the data, hence establishing a connection between existing manifold learning approaches and topological methods. This generalization allows to perform unsupervised feature selection on features that have a complex combinatorial structure (for instance, features that are evaluated in pairs of points, triplets, etc.) as well as to take into account the topological structure spanned by the data at a given scale.

There are at least two directions that we believe deserve further investigation. Graph Laplacians admit elegant interpretations in terms of random walks, Markov chains, and diffusion processes (3). It is still unclear to us how this formalism can be generalized to higher-dimensional combinatorial Laplacians and random walks on simplicial complexes, although some progress has already been done in that direction (23, 24). Additionally, given a filtration of Čech complexes with varying scale $\varepsilon$, it has been shown that its cohomology (spanned by the zero eigenvalue $q$-forms of the combinatorial Laplacian) can be formulated in terms of the theory of persistence (25). It is an open question whether this formulation can be extended to non-zero eigenvalues of the combinatorial Laplacian. Addressing these questions will further contribute to unifying conceptually the tools of manifold learning and topological data analysis.